\documentclass{llncs}

\usepackage{times}
\usepackage{mlapa}
\usepackage{amsfonts}
\usepackage{amsmath}
\usepackage{amssymb}
\usepackage{amstext}
\usepackage{latexsym}
\usepackage{color}
\usepackage{graphics}
\usepackage{enumerate}
\usepackage{url}
\usepackage{bbm}

\def\Nset{\mathbb{N}}

\def\Rset{\mathbb{R}}

\def\C{\mathsf{C}}

\DeclareMathOperator*{\E}{\rm E}

\DeclareMathOperator{\cond}{cond}
\providecommand{\abs}[1]{\lvert#1\rvert}
\providecommand{\norm}[1]{\lVert#1\rVert}

\newcommand{\bi}{\begin{itemize}}
\newcommand{\ei}{\end{itemize}}
\newcommand{\be}{\begin{enumerate}}
\newcommand{\ee}{\end{enumerate}}
\newcommand{\bd}{\begin{description}}
\newcommand{\ed}{\end{description}}
\newcommand{\ignore}[1]{}
\newcommand{\D}{{\mathcal D}}
\newcommand{\W}{{\mathcal W}}
\newcommand{\U}{{\mathcal U}}
\renewcommand{\P}{{\mathcal P}}

\newcommand{\set}[1]{\{#1\}}

\newcommand{\dotp}[2]{\left\langle #1 , #2 \right\rangle}

\newcommand{\e}{\epsilon}
\newcommand{\mat}[1]{{\mathbf #1}}
\newcommand{\K}{\mat{K}}
\newcommand{\iprod}[2]{\left\langle #1 , #2 \right\rangle}

\title{Sample Selection Bias Correction Theory}
\titlerunning{Sample Bias Correction}

\author{Corinna Cortes \inst{1}\! \and
Mehryar Mohri \inst{2, 1}\! \and
Michael Riley \inst{1}\! \and
Afshin Rostamizadeh\inst{2}\thanks{Student's submission, to be considered as a candidate for the E.M. Gold Award.}
}
\authorrunning{Cortes, Mohri, Riley, and Rostamizadeh}

\institute{Google Research, \\
76 Ninth Avenue, New York, NY 10011. \and
  Courant Institute of Mathematical Sciences, \\
  251 Mercer Street, New York, NY 10012.}

\begin{document}
\maketitle

\begin{abstract}
  This paper presents a theoretical analysis of sample selection bias
  correction. The sample bias correction technique commonly used in
  machine learning consists of reweighting the cost of an error on
  each training point of a biased sample to more closely reflect the
  unbiased distribution. This relies on weights derived by various
  estimation techniques based on finite samples. We analyze the effect
  of an error in that estimation on the accuracy of the hypothesis
  returned by the learning algorithm for two estimation
  techniques: a cluster-based estimation technique and kernel mean
  matching.  We also report the results of sample bias correction
  experiments with several data sets using these techniques. Our
  analysis is based on the novel concept of \emph{distributional
    stability} which generalizes the existing concept of point-based
  stability. Much of our work and proof techniques can be used to
  analyze other importance weighting techniques and their effect on
  accuracy when using a distributionally stable algorithm.
\end{abstract}

\section{Introduction}

In the standard formulation of machine learning problems, the learning
algorithm receives training and test samples drawn according to the same
distribution. However, this assumption often does not hold in practice.
The training sample available is \emph{biased} in some way, which may be
due to a variety of practical reasons such as the cost of data labeling
or acquisition. The problem occurs in many areas such as astronomy, 
econometrics, and species habitat modeling.

In a common instance of this problem, points are drawn according to
the test distribution but not all of them are made available to the
learner. This is called the \emph{sample selection bias problem}.
Remarkably, it is often possible to correct this bias by using large
amounts of unlabeled data.

The problem of sample selection bias correction for linear regression
has been extensively studied in econometrics and statistics
\cite{heckman,little_and_rubin} with the pioneering work of
\emcite{heckman}. Several recent machine learning publications
\cite{elkan,zadrozny,zadrozny2,zadrozny3,dudik} have also dealt with
this problem. The main correction technique used in all of these
publications consists of reweighting the cost of training point errors
to more closely reflect that of the test distribution. This is in fact
a technique commonly used in statistics and machine learning for a
variety of problems of this type \cite{little_and_rubin}. With the
exact weights, this reweighting could optimally correct the bias, but,
in practice, the weights are based on an estimate of the sampling
probability from finite data sets. Thus, it is important to determine
to what extent the error in this estimation can affect the accuracy of
the hypothesis returned by the learning algorithm. To our knowledge,
this problem has not been analyzed in a general manner.

This paper gives a theoretical analysis of sample selection bias
correction. Our analysis is based on the novel concept of
\emph{distributional stability} which generalizes the point-based
stability introduced and analyzed by previous authors
\cite{devroye,kearns,bousquet}. We show that large families of
learning algorithms, including all kernel-based regularization
algorithms such as Support Vector Regression (SVR) \cite{vapnik98} or
kernel ridge regression \cite{saunders} are distributionally stable
and we give the expression of their stability coefficient for both the
$l_1$ and $l_2$ distance.

We then analyze two commonly used sample bias correction techniques: a
cluster-based estimation technique and kernel mean matching (KMM)
\cite{huang-nips06}. For each of these techniques, we derive bounds on
the difference of the error rate of the hypothesis returned by a
distributionally stable algorithm when using that estimation technique
versus using perfect reweighting. We briefly discuss and compare these
bounds and also report the results of experiments with both estimation
techniques for several publicly available machine learning data
sets. Much of our work and proof techniques can be used to analyze
other importance weighting techniques and their effect on accuracy
when used in combination with a distributionally stable algorithm.

The remaining sections of this paper are organized as
follows. Section~\ref{sec:Sample Selection Bias Correction} describes
in detail the sample selection bias correction
technique. Section~\ref{sec:Distributional Stability} introduces the
concept of distributional stability and proves the distributional
stability of kernel-based regularization
algorithms. Section~\ref{sec:Analysis of the Effect of Estimation
Error} analyzes the effect of estimation error using distributionally
stable algorithms for both the cluster-based and the KMM estimation
techniques. Section~\ref{sec:Experimental Results} reports the results
of experiments with several data sets comparing these estimation
techniques.

\section{Sample Selection Bias Correction}
\label{sec:Sample Selection Bias Correction}

\subsection{Problem}

Let $X$ denote the input space and $Y$ the label set, which may be
$\set{0, 1}$ in classification or a subset of $\Rset$ in regression
estimation problems, and let $\D$ denote the \emph{true distribution}
over $X \times Y$ according to which test points are drawn.  In the
sample selection bias problem, some pairs $z \!=\! (x, y)$ drawn according
to $\D$ are not made available to the learning algorithm. The learning
algorithm receives a training sample $S$ of $m$ labeled points $z_1,
\ldots, z_m$ drawn according to a \emph{biased distribution} $\D'$
over $X \times Y$. This sample bias can be represented by a random
variable $s$ taking values in $\set{0, 1}$: when $s \!=\!  1$ the
point is sampled, otherwise it is not. Thus, by definition of the
sample selection bias, the support of the biased distribution $\D'$ is
included in that of the true distribution $\D$.

As in standard learning scenarios, the objective of the learning
algorithm is to select a hypothesis $h$ out of a hypothesis set $H$
with a small generalization error $R(h)$ with respect to the true
distribution $\D$, $R(h) = \E_{(x, y) \sim \D}[c(h, z)]$, where $c(h,
z)$ is the cost of the error of $h$ on point $z \in X \times Y$.

While the sample $S$ is collected in some biased manner, it is often
possible to derive some information about the nature of the bias. This
can be done by exploiting large amounts of unlabeled data drawn
according to the true distribution $\D$, which is often available in
practice. Thus, in the following let $U$ be a sample drawn according
to $\D$ and $S \subseteq U$ a labeled but biased sub-sample.

\subsection{Weighted Samples}

A \emph{weighted sample} $S_w$ is a training sample $S$ of $m$ labeled
points, $z_1, \ldots, z_m$ drawn i.i.d.\ from $X \times Y$, that is
augmented with a non-negative weight $w_i \geq 0$ for each point
$z_i$. This weight is used to emphasize or de-emphasize the cost of an
error on $z_i$ as in the so-called \emph{importance weighting} or
\emph{cost-sensitive learning} \cite{elkan,zadrozny2}. One could use
the weights $w_i$ to derive an equivalent unweighted sample $S'$ where
the multiplicity of $z_i$ would reflect its weight $w_i$, but most
learning algorithms, e.g., decision trees, logistic regression,
AdaBoost, Support Vector Machines (SVMs), kernel ridge regression, can
directly accept a weighted sample $S_w$. 
\ignore{
For example, in the case of
AdaBoost, the weights can be normalized to sum to one and used as the
initial weight distribution instead of the uniform distribution, and,
for SVMs, the objective function can be modified by multiplying the
slack variable $\xi_i$ by $w_i$.
}
We will refer to algorithms that can
directly take $S_w$ as input as \emph{weight-sensitive algorithms}.

The empirical error of a hypothesis $h$ on a weighted sample $S_w$ is
defined as
\begin{equation}
\widehat R_w(h) = \sum_{i = 1}^m  w_i \, c(h, z_i).
\end{equation}

\begin{proposition}
\label{prop:risk}
  Let $\D'$ be a distribution whose support coincides with that of
  $\D$ and let $S_w$ be a weighted sample with $w_i =
  \Pr_{\D}(z_i)/\Pr_{\D'}(z_i)$ for all points $z_i$ in $S$. Then,
\begin{equation}
  \E_{S \sim \D'} [\widehat R_w(h)] = R(h) = \E_{z \sim \D} [c(h, z)] .
\end{equation}
\end{proposition}
\begin{proof}
Since the sample points are drawn i.i.d.,
\begin{equation}
  \E_{S \sim \D'} [\widehat R_w(h)] =
  \frac{1}{m} \sum_{z} \E_{S \sim \D'} [w_i c(h, z_i)] = \E_{z_1 \sim \D'} [w_1 c(h, z_1)].
\end{equation}
By definition of $w$ and the fact that the support of $D$ and $D'$
coincide, the right-hand side can be rewritten as follows
\begin{equation}
\sum_{\D'(z_1) \neq 0} {\Pr_{\D}(z_1) \over \Pr_{\D'}(z_1)} \Pr_{\D'}(z_1)\, c(h, z_1)  =
\sum_{\D(z_1) \neq 0} \Pr_{\D}(z_1)\, c(h, z_1) = \E_{z_1 \sim \D} [c(h, z_1)].
\end{equation}
This last term is the definition of the generalization error $R(h)$.\qed
\end{proof}

\subsection{Bias Correction}

The probability of drawing $z = (x, y)$ according to the true but unobserved
distribution $\D$ can be straightforwardly related to the observed distribution
$\D'$.  By definition of the random variable $s$, the observed biased
distribution $\D'$ can be expressed by $\Pr_{\D'}[z] = \Pr_{\D}[z | s = 1]$. We
will assume that all points $z$ in the support of $D$ can be
sampled with a non-zero probability so the support of $\D$ and $\D'$ coincide. Thus for all $z \in X \times Y$, $ \Pr[s =
1 | z] \neq 0$. Then, by the Bayes formula, for all $z$ in the support of $D$,
\begin{equation}
\label{eq:bayes}
\Pr_{\D}[z]  = {\Pr[z | s = 1] \Pr[s = 1] \over \Pr[s = 1 | z]}
 = {\Pr[s = 1] \over \Pr[s = 1 | z]} \Pr_{\D'}[z].
\end{equation}
Thus, if we were given the probabilities $\Pr[s = 1]$ and $\Pr[s = 1 |
z]$, we could derive the true probability $\Pr_{\D}$ from the biased
one $\Pr_{\D'}$ exactly and correct the sample selection bias. 

It is important to note that this correction is only needed for the
training sample $S$, since it is the only source of selection
bias. With a weight-sensitive algorithm, it suffices to reweight each
sample $z_i$ with the weight $w_i = {\Pr[s = 1] \over \Pr[s = 1 |
  z_i]}$. Thus, $\Pr[s = 1 | z]$ need not be estimated for all points
$z$ but only for those falling in the training sample
$S$.\ignore{\footnote{The claim that the sample selection bias
    correction technique just described requires estimating both the
    biased and target distributions is not correct
    \cite{huang-nips06}.  It only requires estimating the sample bias
    probabilities and only for the points of the training data.} }  By
Proposition~\ref{prop:risk}, the expected value of the empirical error
after reweighting is the same as if we were given samples from the
true distribution and the usual generalization bounds hold for
$\widehat R(h)$ and $R(h)$.

When the sampling probability is independent of the labels, as it is
commonly assumed in many settings \singlecite{zadrozny,zadrozny2},
$\Pr[s = 1 | z] = \Pr[s = 1 | x]$, and Equation~\ref{eq:bayes} can be
re-written as
\begin{equation}
\label{eq:nbayes}
  \Pr_{\D}[z] = {\Pr[s = 1] \over \Pr[s = 1 | x]} \Pr_{\D'}[z].
\end{equation}
In that case, the probabilities $\Pr[s = 1]$ and $\Pr[s = 1 | x]$
needed to reconstitute $\Pr_{\D}$ from $\Pr_{\D'}$ do not depend on
the labels and thus can be estimated using the unlabeled points in
$U$. Moreover, as already mentioned, for weight-sensitive algorithms,
it suffices to estimate $\Pr[s = 1 | x_i]$ for the points $x_i$ 
of the training data; no generalization is needed.

A simple case is when the points are defined over a discrete
set.\footnote{This can be as a result of a quantization or clustering
  technique as discussed later.}  $\Pr[s = 1 | x]$ can then be
estimated from the frequency $m_x / n_x$, where $m_x$ denotes the
number of times $x$ appeared in $S \subseteq U$ and $n_x$ the number
of times $x$ appeared in the full data set $U$. \ignore{ As already
  pointed out, no generalization is needed here since the
  probabilities need to be estimated only for the points appearing in
  the training data.  } $\Pr[s = 1]$ can be estimated by the quantity
$|S|/|U|$. However, since $\Pr[s = 1]$ is a constant independent of
$x$, its estimation is not even necessary.

\ignore{
Another case of interest is when the sampling probability only depends
on the labels $y$: $\Pr[s = 1 | z] = \Pr[s = 1 | y]$. As in the
previous case, sample selection bias can be corrected via a
reweighting based on $1/\Pr[s = 1 | y]$ provided that sufficient
additional labeled data besides $S$ is available. The most general
case, which requires estimating $\Pr[s = 1 | z]$, also requires
sufficient amounts of additional labeled data which is often very
limited in applications.
}
 
\ignore{
Other scenarios of interest include the case where the sampling
probability depends only on the labels, $\Pr[s = 1 | z] = \Pr[s = 1 |
y]$, or, in the most general case, on the pair $(x,y)$. These
scenarios require additional labeled data which is often very limited
in applications.

In what follows, we will be considering sample selection bias
correction in the common case where $\Pr[s = 1 | z] = \Pr[s = 1 | x]$.}

If the estimation of the sampling probability $\Pr[s = 1 | x]$ from
the unlabeled data set $U$ were exact, then the reweighting just
discussed could correct the sample bias optimally. Several techniques
have been commonly used to estimate the reweighting quantities.  But,
these estimate weights are not guaranteed to be exact. 
The next section addresses how the error in that estimation affects
the error rate of the hypothesis returned by the learning algorithm.

\section{Distributional Stability}
\label{sec:Distributional Stability}

Here, we will examine the effect on the error of the hypothesis
returned by the learning algorithm in response to a change in the way
the training points are weighted. Since the weights are non-negative,
we can assume that they are normalized and define a distribution over
the training sample.  This study can be viewed as a generalization of
stability analysis where a single sample point is changed
\cite{devroye,kearns,bousquet} to the more general case of
\emph{distributional stability} where the sample's weight distribution
is changed.

Thus, in this section the sample weight $\W$ of $S_{\W}$ defines a
distribution over $S$. For a fixed learning algorithm $L$ and a fixed
sample $S$, we will denote by $h_\W$ the hypothesis returned by $L$
for the weighted sample $S_\W$. We will denote by $d(\W, \W')$ a
divergence measure for two distributions $\W$ and $\W'$. There are
many standard measures for the divergences or distances between two
distributions, including the relative entropy, the Hellinger distance,
and the $l_p$ distance.

\begin{definition}[Distributional $\beta$-Stability]
  A learning algorithm $L$ is said to be
  \emph{distributionally $\beta$-stable} for the divergence measure
  $d$ if for any two weighted samples $S_\W$ and $S_{\W'}$, 
\begin{equation}
  \forall z \in X \times Y, \quad \abs{c(h_\W, z) - c(h_{\W'}, z)} \leq \beta\, d(\W, \W').
\end{equation}
\end{definition}
Thus, an algorithm is distributionally stable when small changes to a
weighted sample's distribution, as measured by a divergence $d$,
result in a small change in the cost of an error at any point.
The following proposition follows directly from the definition of 
distributional stability.
\begin{proposition}
\label{prop:prop1}
  Let $L$ be a distributionally $\beta$-stable algorithm and let $h_\W$
  ($h_{\W'}$) denote the hypothesis returned by $L$ when trained on the
  weighted sample $S_\W$ (resp. $S_{\W'}$). Let $\W_T$ denote the
  distribution according to which test points are drawn. Then, the
  following holds
\begin{equation}
\abs{R(h_\W) - R(h_{\W'})} \leq \beta\, d(\W, \W').
\end{equation}
\end{proposition}
\begin{proof}
By the distributional stability of the algorithm,
\begin{equation}
  \E_{z \sim \W_T}[\abs{c(z, h_\W) - c(z, h_{\W'})}] \leq \beta\, d(\W, \W'),
\end{equation}
which implies the statement of the proposition.\qed
\end{proof}

\subsection{Distributional Stability of Kernel-Based Regularization Algorithms}

Here, we show that kernel-based regularization algorithms are
distributionally $\beta$-stable. This family of algorithms includes,
among others, Support Vector Regression (SVR) and
kernel ridge regression. Other algorithms such as
those based on the relative entropy regularization can be shown to be
distributionally $\beta$-stable in a similar way as for point-based
stability. Our results also apply to classification algorithms such as
Support Vector Machine (SVM) \cite{cortes&vapnik} using a margin-based
loss function $l_\gamma$ as in \cite{bousquet}.

We will assume that the cost function $c$ is
\emph{$\sigma$-admissible}, that is there exists
$\sigma \in \Rset_+$ such that for any two hypotheses $h, h' \in H$
and for all $z = (x, y) \in X \times Y$,
\begin{equation}
|c(h, z) - c(h', z)| \leq \sigma |h(x) - h'(x)|.
\end{equation}
This assumption holds for the quadratic cost and most other cost
functions when the hypothesis set and the set of output labels are
bounded by some $M \in \Rset_+$: $\forall h \in H, \forall x \in X,
|h(x)| \leq M$ and $\forall y \in Y, |y| \leq M$. We will also assume
that $c$ is differentiable. This assumption is in fact not necessary
and all of our results hold without it, but it makes the presentation
simpler.

Let $N\colon H \to \Rset_+$ be a function defined over the hypothesis set.
Regularization-based algorithms minimize an objective of the form
\begin{equation}
F_\W(h) = \widehat R_\W(h) + \lambda N(h),
\end{equation}
where $\lambda \geq 0$ is a trade-off parameter. We denote by $B_F$
the Bregman divergence associated to a convex function $F$, $B_F(f
\Arrowvert g) = F(f) - F(g) - \dotp{f - g}{\nabla F(g)}$, and define
$\Delta h$ as $\Delta h = h' - h$.

\begin{lemma}
\label{lemma:l1-dist}
Let the hypothesis set $H$ be a vector space. Assume that $N$ is a
proper closed convex function and that $N$ is differentiable.
Assume that $F_\W$ admits a minimizer $h \in H$ and $F_{\W'}$ a
minimizer $h' \in H$. Then, the following bound holds,
\begin{equation}
B_N(h' \Arrowvert h) + B_N(h \Arrowvert h') \leq {\sigma\, l_1(\W, \W') \over \lambda} 
  \sup_{x \in S} |\Delta h(x)|.
\end{equation}
\end{lemma}
\begin{proof}
Since $B_{F_\W} = B_{\widehat R_\W} + \lambda B_{N}$ and $B_{F_{\W'}} =
B_{\widehat R_{\W'}} + \lambda B_{N}$, and a Bregman divergence is
non-negative,
$\lambda \bigl(B_N(h' \Arrowvert h) + B_N(h \Arrowvert h')\bigr) \leq B_{F_\W}(h' \Arrowvert h) + B_{F_{\W'}}(h \Arrowvert h')$.
By the definition of $h$ and $h'$ as the minimizers of $F_\W$ and
$F_{\W'}$,
\begin{equation}
  B_{F_\W}(h' \Arrowvert h) + B_{F_{\W'}}(h \Arrowvert h') = \widehat
  R_{F_\W}(h') - \widehat R_{F_\W}(h) + \widehat R_{F_{\W'}}(h) -
  \widehat R_{F_{\W'}}(h').
\end{equation}
Thus, by the $\sigma$-admissibility of the cost function $c$, using
the notation $\W_i = \W(x_i)$ and $\W'_i = \W'(x_i)$,
\begin{equation}
{\begin{split}
& \lambda \bigl(B_N(h' \Arrowvert h) + B_N(h \Arrowvert h')\bigr)
\leq \widehat R_{F_\W}(h') - \widehat R_{F_\W}(h) + \widehat R_{F_{\W'}}(h) - \widehat R_{F_{\W'}}(h') \\
& = \sum_{i=1}^m \biggl[ c(h',z_i)\W_i - c(h,z_i)\W_i + c(h,z_i)\W'_i - c(h',z_i)\W'_i \biggr] \\
& = \sum_{i=1}^m \biggl[ (c(h',z_i) - c(h,z_i))(\W_i - \W'_i) \biggr] 
= \sum_{i=1}^m \biggl[ \sigma |\Delta h(x_i)|(\W_i - \W'_i) \biggr] \\
& \leq \sigma  l_1(\W, \W') \sup_{x \in S} |\Delta h(x)|,
\end{split}}
\end{equation}
which establishes the lemma.\qed
\end{proof}
Given $x_1, \ldots, x_m \in X$ and a positive definite symmetric (PDS)
kernel $K$, we denote by $\K \in \Rset^{m \times m}$ the kernel matrix
defined by $\K_{ij} = K(x_i, x_j)$ and by $\lambda_{\max}(\K) \in
\Rset_+$ the largest eigenvalue of $\K$.

\begin{lemma}
\label{lemma:l2-dist}
Let $H$ be a reproducing kernel Hilbert space with kernel $K$ and let
the regularization function $N$ be defined by $N(\cdot) =
\norm{\cdot}_K^2$. Then, the following bound holds,
\begin{equation}
  B_N(h' \Arrowvert h) + B_N(h \Arrowvert h') \leq {\sigma \lambda_{\max}^{\frac 1 2}(\K) \, l_2(\W, \W') \over \lambda} 
  \norm{\Delta h}_2.
\end{equation}
\end{lemma}
\begin{proof}
As in the proof of Lemma~\ref{lemma:l1-dist},
\begin{equation}
  \lambda \bigl(B_N(h' \Arrowvert h) + B_N(h \Arrowvert h')\bigr)
  \leq
  \sum_{i=1}^m \biggl[ (c(h',z_i) - c(h,z_i))(\W_i - \W'_i) \biggr].
\end{equation}
By definition of a reproducing kernel Hilbert space $H$, for any
hypothesis $h \in H$, $\forall x \in X, h(x) = \dotp{h}{K(x, \cdot)}$
and thus also for any $\Delta h = h' - h$ with $h, h' \in H$, $\forall
x \in X, \Delta h(x) = \dotp{\Delta h}{K(x, \cdot)}$. 
Let $\Delta \W_i$ denote $\W'_i - \W_i$, $\Delta \W$ the
vector whose components are the $\Delta \W_i$'s, and let $V$
denote $B_N(h' \Arrowvert h) + B_N(h \Arrowvert h')$.  Using
$\sigma$-admissibility,
$V  \leq \sigma \sum_{i = 1}^m |\Delta h(x_i) \, \Delta \W_i| = \sigma \sum_{i = 1}^m |\dotp{\Delta h}{\Delta \W_i K(x_i, \cdot)}|$.
Let $\e_i \in \set{-1, +1}$ denote the sign of $\dotp{\Delta h}{\Delta
  \W_i K(x_i, \cdot)}$. Then,
\begin{equation}
{\begin{split}
& V \leq \sigma \dotp{\Delta h}{\sum_{i = 1}^m \e_i \Delta \W_i K(x_i, \cdot)} 
\leq \sigma \norm{\Delta h}_K\, \norm{{\sum_{i = 1}^m \e_i\Delta \W_i K(x_i, \cdot)}}_K \\
& = \sigma \norm{\Delta h}_K \bigl({\sum_{i,j = 1}^m \e_i\e_j \Delta \W_i\Delta \W_j K(x_i, x_j)}\bigr)^{1/2} \\
& = \sigma \norm{\Delta h}_K \bigl[\Delta (\W\e)^\top  \K \Delta (\W \e) \bigr]^{\frac 1 2} 
\leq \sigma \norm{\Delta h}_K \norm{\Delta \W}_2 \lambda_{\max}^{\frac 1 2}(\K).
\end{split}}
\end{equation}
In this derivation, the second inequality follows from the
Cauchy-Schwarz inequality and the last inequality from the standard
property of the Rayleigh quotient for PDS matrices. Since
$\norm{\Delta \W}_2 = l_2(\W, \W')$, this proves the lemma.\qed
\end{proof}

\begin{theorem}
\label{th:beta_bounds}
Let $K$ be a kernel such that $K(x, x) \leq \kappa < \infty$ for all
$x \in X$. Then, the regularization algorithm based on $N(\cdot) =
\norm{\cdot}_K^2$ is distributionally $\beta$-stable for the $l_1$
distance with $\beta \leq \frac {\sigma^2
  \kappa^2}{2 \lambda}$, and for the $l_2$ distance with $\beta \leq
\frac {\sigma^2 \kappa \lambda_{\max}^{1 \over 2}(\K)}{2 \lambda}$.
\end{theorem}
\begin{proof}
  \ignore{The proof of the theorem is similar to that of Theorem 22 by
    \emcite{bousquet} but with the use of our
    Lemmas~\ref{lemma:l1-dist}-\ref{lemma:l2-dist} instead.  The
    details are given to make this presentation self-contained.}
  For $N(\cdot) = \norm{\cdot}_K^2$, we have
  $B_N(h' \Arrowvert h) = \norm{h' - h}_K^2$, thus $B_N(h' \Arrowvert
  h) + B_N(h \Arrowvert h') = 2 \norm{\Delta h}_K^2$ and by
  Lemma~\ref{lemma:l1-dist}, 
\begin{equation}
      2 \norm{\Delta h}_K^2 \leq {\sigma\,
        l_1(\W, \W') \over \lambda} \sup_{x \in S} |\Delta h(x)|
      \leq {\sigma\, l_1(\W, \W') \over \lambda} \kappa ||\Delta h||_K.
\end{equation}
Thus $\norm{\Delta h}_K \leq {\sigma \kappa \, l_1(\W, \W') \over
  2 \lambda}$. By $\sigma$-admissibility of $c$,
\begin{equation}
 \forall z \in X \times Y, \abs{c(h', z) - c(h, z)} \leq \sigma \abs{\Delta h (x)} \leq \kappa \sigma \norm{\Delta h}_K.
\end{equation}
Therefore, 
\begin{equation}
\forall z \in X \times Y, \abs{c(h', z) - c(h, z)} \leq {\sigma^2 \kappa^2 \, l_1(\W, \W') \over 2 \lambda},
\end{equation}
which shows the distributional stability of a kernel-based
regularization algorithm for the $l_1$ distance. Using
Lemma~\ref{lemma:l2-dist}, a similar derivation leads to
\begin{equation}
 \forall z \in X \times Y, \abs{c(h', z) - c(h, z)} \leq {\sigma^2 \kappa \lambda_{\max}^{1 \over 2}(\K) \, l_2(\W, \W') \over 2 \lambda},
\end{equation}
which shows the distributional stability of a kernel-based
regularization algorithm for the $l_2$ distance.\qed
\end{proof}
Note that the standard setting of a sample with no weight is
equivalent to a weighted sample with the uniform distribution $\W_{\U}$:
each point is assigned the weight $1/m$. Removing a single
point, say $x_1$, is equivalent to assigning weight $0$ to $x_1$
and $1/(m - 1)$ to others. Let $\W_{\U'}$ be the corresponding distribution,
then
\begin{equation}
  l_1(\W_{\U}, \W_{\U'}) = \frac 1 m + \sum_{i = 1}^{m - 1} \biggl|\frac 1 m -
  \frac {1}{m - 1}\biggr| = \frac 2 m.
\end{equation}
Thus, in the case of kernel-based regularized algorithms and for the
$l_1$ distance, standard uniform $\beta$-stability is a special case
of distributional $\beta$-stability. It can be shown similarly that
$l_2(\W_{\U}, \W_{\U'}) = \frac {1}{\sqrt{m (m -1)}}$.

\section{Effect of Estimation Error for Kernel-Based Regularization Algorithms}
\label{sec:Analysis of the Effect of Estimation Error}

This section analyzes the effect of an error in the estimation of the
weight of a training example on the generalization error of the
hypothesis $h$ returned by a weight-sensitive learning algorithm. We
will examine two estimation techniques: a straightforward
histogram-based or cluster-based method, and kernel mean matching
(KMM) \cite{huang-nips06}.

\subsection{Cluster-Based Estimation}
\label{sec:cluster}

A straightforward estimate of the probability of sampling is based on
the observed empirical frequencies. The ratio of the number of times a
point $x$ appears in $S$ and the number of times it appears in $U$ is
an empirical estimate of $\Pr[s = 1 | x]$. Note that generalization to
unseen points $x$ is not needed since reweighting requires only
assigning weights to the seen training points. However, in general,
training instances are typically unique or very infrequent since
features are real-valued numbers. Instead, features can be discretized
based on a partitioning of the input space $X$. The partitioning may
be based on a simple histogram buckets or the result of a clustering
technique.\ignore{ The assumption is that the sampling probability is
  the same within each partition, that is $\Pr[s = 1 | x] = \Pr[s =
  1 | x']$ for two points $x$ and $x'$ in the same partition.} The
analysis of this section assumes such a prior partitioning of $X$.\ignore{
  Thus, the points in $S$ and $U$ may correspond to clusters.}

We shall analyze how fast the resulting empirical frequencies converge
to the true sampling probability. For $x \in U$, let $U_x$ denote the
subsample of $U$ containing exactly all the instances of $x$ and let
$n = |U|$ and $n_x = |U_x|$. Furthermore, let $n'$ denote the number
of unique points in the sample $U$. Similarly, we define $S_x$, $m$,
$m_x$ and $m'$ for the set $S$. Additionally, denote by $p_0 = \min_{x
  \in U} \Pr[x] \neq 0$. \begin{lemma} \label{lem:emp_convergence} Let
  $\delta > 0$. Then, with probability at least $1 - \delta$, the
  following inequality holds for all $x$ in $S$: 
\begin{equation}
  \Bigl|\Pr[s = 1 | x] - \frac{m_x}{n_x} \Bigr| \leq
  \sqrt{\frac {\log 2m' + \log {1 \over \delta }}{p_0n}}.
\end{equation}
\end{lemma}

\begin{proof}
For a fixed $x \in U$, by Hoeffding's inequality,
\begin{equation*}
\small {\begin{split}
\Pr_{U}\Bigl[\bigl|\Pr[s = 1 | x] - \frac{m_x}{n_x}\bigr| \geq \e\Bigr] 
& =  \sum_{i=1}^n \Pr_{x}\Bigl[|\Pr[s = 1 | x] - \frac{m_x}{i}| \geq \e \mid n_x = i\Bigr] \Pr[n_x = i]\\
& \leq \sum_{i=1}^n 2 e^{-2i\e^2} \Pr_U[n_x = i].
\end{split}}
\end{equation*}
Since $n_x$ is a binomial random variable with parameters $\Pr_U[x] =
p_x$ and $n$, this last term can be expressed more explicitly and
bounded as follows:
\begin{equation*}
\small {\begin{split}
      2 \sum_{i=1}^n e^{-2i\e^2} \Pr_U[n_x = i] 
      & \leq 2 \sum_{i = 0}^n e^{-2i\e^2} \binom{n}{i} p_x^i (1 - p_x)^{n-i}
      = 2 (p_xe^{-2\e^2} + (1 - p_x) )^n \\
    & = 2 (1 - p_x (1 - e^{-2 \e^2}))^n
    \leq 2 \exp(-p_x n (1 - e^{-2 \e^2})).
\end{split}}
\end{equation*}
Since for $x \in [0, 1]$, $1 - e^{-x} \geq x/2$, this shows that
for $\e \in [0, 1]$,
\begin{equation}
  \small  \Pr_{U}\Bigl[\bigl|\Pr[s = 1 | x] - \frac{m_x}{n_x}\bigr| \geq \e\Bigr]
  \leq 2 e^{-p_x n \e^2}.
\end{equation}
By the union bound and the definition of $p_0$,
\begin{equation*}
  \small \Pr_{U}\Bigl[\exists x \in S: \bigl|\Pr[s = 1 | x] - 
  \frac{m_x}{n_x}\bigr| \geq \e\Bigr]  \leq 2 m' e^{-p_0 n \e^2}.
\end{equation*}
Setting $\delta$ to match the upper bound yields the 
statement of the lemma.\qed
\end{proof}
The following proposition bounds the distance between the distribution
$\W$ corresponding to a perfectly reweighted sample ($S_\W$) and the
one corresponding to a sample that is reweighted according to the
observed bias ($S_{\widehat \W}$). For a sampled point $x_i = x$, these
distributions are defined as follows:
\begin{equation}
\W(x_i) = {1 \over m} {1 \over p(x_i)}
\quad \text{and} \quad
\widehat \W(x_i) = {1 \over m} {1 \over \hat p(x_i)},
\end{equation}
where, for a \emph{distinct} point $x$ equal to the \emph{sampled} point
$x_i$, we define $p(x_i) = \Pr[s = 1 | x]$ and $\hat p(x_i) =
\frac{m_x}{n_x}$.

\begin{proposition}
  Let $B = \displaystyle \max_{i = 1, \ldots, m} \max(1/p(x_i), 1/ \hat
    p(x_i))$.  Then, the $l_1$ and $l_2$ distances of the
  distributions $\W$ and $\widehat \W$ can be bounded as follows,
\begin{equation}
{\begin{split}
l_1(\W, \widehat \W) \leq B^2 \sqrt{\frac {\log 2m' + \log {1 \over \delta }}{p_0n}} 
\text{ and } l_2(\W, \widehat \W) \leq B^2 \sqrt{\frac {\log 2m' + \log {1 \over \delta }}{p_0nm}}.
\end{split}}
\end{equation}
\end{proposition}

\begin{proof}
By definition of the $l_2$ distance, 
\begin{equation*}
{\begin{split}
l_2^2(\W, \widehat \W) 
& = {1 \over m^2} \sum_{i=1}^m \left( {1 \over p(x_i)} - {1 \over
\hat p(x_i)} \right)^2 
 = {1 \over m^2} \sum_{i=1}^m \left( {p(x_i) - \hat p(x_i) \over p(x_i)
\hat p(x_i)} \right)^2 \\
& \leq {B^4 \over m} \max_i ( p(x_i) - \hat p(x_i) )^2.
\end{split}}
\end{equation*}
It can be shown similarly that $\small{l_1(\W, \widehat \W) \leq B^2
  \max_i \left| p(x_i) - \hat p(x_i) \right|}$. The application of the
uniform convergence bound of Lemma~\ref{lem:emp_convergence} directly
yields the statement of the proposition.\qed
\end{proof}
The following theorem provides a bound on the difference between the
generalization error of the hypothesis returned by a kernel-based
regularization algorithm when trained on the perfectly unbiased
distribution, and the one trained on the sample bias-corrected using
frequency estimates.
\begin{theorem}
\label{th:main}
Let $K$ be a PDS kernel such that $K(x, x) \leq \kappa \!<\! \infty$
for all $x \in X$. Let $h_\W$ be the hypothesis returned by the
regularization algorithm based on $N(\cdot) = \norm{\cdot}_K^2$ using
$S_\W$, and $h_{\widehat \W}$ the one returned after training the same
algorithm on $S_{\widehat \W}$. Then, for any $\delta > 0$, with
probability at least $1 - \delta$, the difference in generalization
error of these hypotheses is bounded as follows
\begin{equation}
{\begin{split}
& |R(h_\W) - R(h_{\widehat \W})| \leq \dfrac {\sigma^2 \kappa^2
    B^2}{2
    \lambda} \sqrt{\dfrac {\log 2m' + \log {1 \over \delta }}{p_0n}}\\
& |R(h_\W) - R(h_{\widehat \W})| \leq \dfrac {\sigma^2 \kappa
    \lambda_{\max}^{1 \over 2}(\K) B^2}{2 \lambda}\sqrt{\dfrac {\log
      2m' + \log {1 \over \delta }}{p_0nm}}.
\end{split}}
\end{equation}
\end{theorem}
\begin{proof}
  The result follows from Proposition~\ref{prop:prop1}, the distributional
  stability and the bounds on the stability coefficient $\beta$ for
  kernel-based regularization algorithms
  (Theorem~\ref{th:beta_bounds}), and the bounds on the $l_1$ and
  $l_2$ distances between the correct distribution $\W$ and the
  estimate $\widehat \W$.\qed
\end{proof}
Let $n_0$ be the number of occurrences, in $U$, of the least frequent
training example. For large enough $n$, $p_0 n \approx n_0$, thus the
theorem suggests that the difference of error rate between the
hypothesis returned after an optimal reweighting versus the one based
on frequency estimates goes to zero as $\sqrt{\frac{\log m'}{n_0}}$. In
practice, $m' \leq m$, the number of distinct points in $S$ is small,
a fortiori, $\log m'$ is very small, thus, the convergence rate
depends essentially on the rate at which $n_0$ increases. Additionally, if
$\lambda_{\max}(K) \leq m$ (such as with Gaussian kernels), the $l_2$-based
bound will provide convergence that is at least as fast.

\ignore{
There is a trade-off in the prior quantization or clustering step
between reducing the bound of the theorem to guarantee an error closer
to that which is obtained with an ideal bias correction, and
preserving enough effective sample points to ensure good performance.
Our bound can help evaluate different trade-offs and select
the best.}

\subsection{Kernel Mean Matching}

The following definitions introduced by \emcite{steinwart-jmlr02} will
be needed for the presentation and discussion of the kernel mean
matching (KMM) technique. Let $X$ be a compact metric space and let
$C(X)$ denote the space of all continuous functions over $X$ equipped
with the standard infinite norm $\| \cdot \|_\infty$. Let $K\colon X
\times X \to \Rset$ be a PDS kernel. There
exists a Hilbert space $F$ and a map $\Phi\colon X \to F$ such that
for all $x, y \in X$, $K(x, y) = \iprod{\Phi(x)}{\Phi(y)}$. Note that
for a given kernel $K$, $F$ and $\Phi$ are not unique and that, for
these definitions, $F$ does not need to be a reproducing kernel
Hilbert space (RKHS).

Let $\cal P$ denote the set of all probability distributions over 
$X$ and let $\mu\colon \P \to F$ be the function defined by
\begin{equation}
\forall p \in \P, \quad \mu(p) = \E_{x \sim p} [\Phi(x)].
\end{equation}
A function $g\colon X \to \Rset$ is said to be \emph{induced} by $K$
if there exists $w \in F$ such that for all $x \in X$, $g(x) =
\iprod{w}{\Phi(x)}$. $K$ is said to be \emph{universal} if it is
continuous and if the set of functions induced by $K$ are dense in
$C(X)$.

\begin{theorem}[\emcite{huang-report06}]
\label{th:kmm}
  Let $F$ be a separable Hilbert space and let $K$ be a universal
  kernel with feature space $F$ and feature map $\Phi\colon X \to
  F$. Then, $\mu$ is injective.
\end{theorem}
\begin{proof} 
The proof given by \emcite{huang-report06} does not seem to be complete, we
have included a complete proof in the Appendix.
\ignore{ The
    authors claim that $\mu$ is a linear function, but its domain of
    definition, $\P$, is not a vector space. Several of the equalities
    in the proof (page 6) are not justified and seem to be
    incorrect.}\qed
\end{proof}
The KMM technique is applicable when the learning algorithm is based
on a universal kernel. The theorem shows that for a universal kernel,
the expected value of the feature vectors induced uniquely determines
the probability distribution. KMM uses this property to reweight
training points so that the average value of the feature vectors for
the training data matches that of the feature vectors for a set of
unlabeled points drawn from the unbiased distribution. 

Let $\gamma_i$ denote the perfect reweighting of the sample point
$x_i$ and $\widehat \gamma_i$ the estimate derived by KMM. Let $B'$
denote the largest possible reweighting coefficient $\gamma$ and let
$\e$ be a positive real number.  We will assume that $\e$ is chosen so
that $\e \leq 1/2$. Then, the following is the KMM constraint
optimization
\begin{equation}
{\begin{split}
& \min_{\gamma} \ G(\gamma) = \|\frac{1}{m} \sum_{i = 1}^m \gamma_i \Phi(x_i) -  \frac{1}{n} \sum_{i = 1}^{n} \Phi(x'_i)\|\\
& \text{subject to} \ \gamma_i \in [0, B'] \wedge \bigl| \frac{1}{m}\sum_{i = 1}^m \gamma_i - 1\bigr| \leq \e.
\end{split}}
\end{equation}
Let $\widehat \gamma$ be the solution of this optimization problem,
then $\frac{1}{m}\sum_{i = 1}^m \widehat \gamma_i = 1 + \e'$ with $-\e \leq \e'
\leq \e$. For $i \in [1, m]$, let $\widehat \gamma'_i = \widehat
\gamma_i/(1 + \e')$. The normalized weights used in KMM's reweighting of the sample are thus defined by $\widehat
\gamma'_i/m$ with $\frac{1}{m}\sum_{i = 1}^m \gamma'_i = 1$.

As in the previous section, given $x_1, \ldots, x_m \in X$ and a
strictly positive definite universal kernel $K$, we denote by $\K \in
\Rset^{m \times m}$ the kernel matrix defined by $\K_{ij} = K(x_i,
x_j)$ and by $\lambda_{\min}(\K) > 0$ the smallest eigenvalue of $\K$.
We also denote by $\cond(\K)$ the condition number of the matrix $\K$:
$\cond(\K) = \lambda_{\max}(\K) / \lambda_{\min}(\K) $.  When $K$ is
universal, it is continuous over the compact $X \times X$ and thus
bounded and there exists $\kappa < \infty$ such that $K(x, x) \leq
\kappa$ for all $x \in X$.

\begin{proposition}
\label{prop:kmm_l2}
  Let $K$ be a strictly positive definite universal kernel. Then, for
  any $\delta > 0$, with probability at least $1 - \delta$, the $l_2$
  distance of the distributions $\widehat \gamma'/m$ and $\gamma/m$ is
  bounded as follows:
\begin{equation}
  \frac{1}{m} \|(\widehat \gamma' - \gamma)\|_2 \leq \frac{2 \e B'
}{\sqrt{m}}  +  \frac{2 \kappa^{1 \over 2}}{\lambda^{1 \over 2}_{\min}(\K)}
\sqrt{\frac{B'^2}{m} + \frac{1}{n}} \biggl(1 + \sqrt{2 \log \frac{2}{\delta}}\biggr).
\end{equation}

\end{proposition}
\begin{proof}
  Since the optimal reweighting $\gamma$ verifies the constraints of the
  optimization, by definition of $\widehat \gamma$ as a minimizer,
  $G(\widehat \gamma) \leq G(\gamma)$. Thus, by the triangle
  inequality,
\begin{equation}
\label{eq:gammas}
\|\frac{1}{m} \sum_{i = 1}^m \widehat \gamma_i \Phi(x_i) -  \frac{1}{m} \sum_{i = 1}^m \gamma_i \Phi(x_i)\| \leq G(\widehat \gamma) + G(\gamma) \leq 2 G(\gamma).
\end{equation}
Let $L$ denote the left-hand side of this inequality: $L = \frac{1}{m}
\|\sum_{i = 1}^m (\widehat \gamma_i - \gamma_i) \Phi(x_i) \|$.  By
definition of the norm in the Hilbert space, $L = \frac{1}{m}
\sqrt{(\widehat \gamma - \gamma)^\top \K (\widehat \gamma -
  \gamma)}$. Then, by the standard bounds for the Rayleigh
quotient of PDS matrices, $L \geq \frac{1}{m}
\lambda^{1 \over 2}_{\min}(\K) \|(\widehat \gamma - \gamma)\|_2$.
This combined with Inequality~\ref{eq:gammas} yields
\begin{equation}
\frac{1}{m} \|(\widehat \gamma - \gamma)\|_2 \leq \frac{2G(\gamma)}{\lambda^{1 \over 2}_{\min}(\K)}.
\end{equation}
Thus, by the triangle inequality,
\begin{equation}
\label{eq:triangle}
{\begin{split}
\frac{1}{m} \|(\widehat \gamma' - \gamma)\|_2 & \leq  \frac{1}{m} \|(\widehat \gamma' - \widehat \gamma)\|_2 + \frac{1}{m} \|(\widehat \gamma - \gamma)\|_2\\
& \leq \frac{|\e'|/m}{1 + \e'} \|\gamma\|_2 + \frac{2G(\gamma)}{\lambda^{1 \over 2}_{\min}(\K)}\\
& \leq \frac{2 |\e'| B' \sqrt{m}}{m}  + \frac{2G(\gamma)}{\lambda^{1 \over 2}_{\min}(\K)}  \leq
\frac{2 \e B' }{\sqrt{m}}  + \frac{2G(\gamma)}{\lambda^{1 \over 2}_{\min}(\K)}.
\end{split}}
\end{equation}
It is not difficult to show using McDiarmid's inequality that for any
$\delta > 0$, with probability at least $1 - \delta$, the following
holds (Lemma 4, \cite{huang-report06}):
\begin{equation}
  G(\gamma) \leq \kappa^{1 \over 2} \sqrt{\frac{B'^2}{m} + \frac{1}{n}} \biggl(1 + \sqrt{2 \log \frac{2}{\delta}}\biggr).
\end{equation}
This combined with Inequality~\ref{eq:triangle} yields
the statement of the proposition.\qed
\end{proof}
The following theorem provides a bound on the difference between the
generalization error of the hypothesis returned by a kernel-based
regularization algorithm when trained on the true distribution, and
the one trained on the sample bias-corrected KMM.
\begin{theorem}
\label{th:main_kmm}
Let $K$ be a strictly positive definite symmetric universal
kernel. Let $h_{\gamma}$ be the hypothesis returned by the
regularization algorithm based on $N(\cdot) = \norm{\cdot}_K^2$ using
$S_{\gamma/m}$ and $h_{\widehat \gamma'}$ the one returned after
training the same algorithm on $S_{\widehat \gamma'/m}$. Then, for any
$\delta > 0$, with probability at least $1 - \delta$, the difference
in generalization error of these hypotheses is bounded as follows
\begin{equation*}
  |R(h_{\gamma}) - R(h_{\widehat {\gamma'}})| \leq 
  \dfrac {\sigma^2 \kappa
    \lambda_{\max}^{1 \over 2}(\K)}{\lambda}
  \left (\frac{\e B' }{\sqrt{m}}  +  \frac{\kappa^{1 \over 2}}{\lambda^{1
\over 2}_{\min}(\K)} \sqrt{\frac{B'^2}{m} + \frac{1}{n}} \biggl(1 + \sqrt{2 \log \frac{2}{\delta}}\biggr)\right).
\end{equation*}
For $\e = 0$, the bound becomes
\begin{equation}
  |R(h_{\gamma}) - R(h_{\widehat {\gamma'}})| \leq 
    \dfrac{\sigma^2 \kappa^{3 \over 2} {\cond}^{1 \over
2}(\K)}{\lambda}\sqrt{\frac{B'^2}{m} + \frac{1}{n}} \biggl(1 + \sqrt{2 \log \frac{2}{\delta}}\biggr).
\end{equation}
\end{theorem}
\begin{proof}
The result follows from Proposition~\ref{prop:prop1} and the bound
of Proposition~\ref{prop:kmm_l2}.\qed
\end{proof}
Comparing this bound for $\e = 0$ with the $l_2$ bound of
Theorem~\ref{th:main_kmm}, we first note that $B$ and $B'$ are
essentially related modulo the constant $\Pr[s = 1]$ which is not
included in the cluster-based reweighting.  Thus, the cluster-based
convergence is of the order $O (\lambda_{\max}^{\frac{1}{2}}(\K) B^2
\sqrt{\frac{\log m'}{p_0 n m}})$ and the KMM convergence of the order
$O( \cond^{\frac{1}{2}}(\K) \frac{B}{\sqrt{m}})$. Taking the ratio of
the former over the latter and noticing $p_0^{-1} \approx O(B)$, we
obtain the expression $O \bigg( \sqrt{\frac{ \lambda_{\min}(\K) B \log
    m'}{n}} \bigg)$. Thus, for $n > \lambda_{min}(\K) B \log(m')$ the
convergence of the cluster-based bound is more favorable, while for
other values the KMM bound converges faster.

\section{Experimental Results}
\label{sec:Experimental Results}

In this section, we will compare the performance of the cluster-based
reweighting technique and the KMM technique empirically. We will first
discuss and analyze the properties of the clustering method and our
particular implementation.

The analysis of Section~\ref{sec:cluster} deals with discrete points
possibly resulting from the use of a quantization or clustering
technique. However, due to the relatively small size of the public
training sets available, clustering could leave us with few cluster
representatives to train with.  Instead, in our experiments, we only
used the clusters to estimate sampling probabilities and applied these
weights to the full set of training points. As the following
proposition shows, the $l_1$ and $l_2$ distance bounds of Proposition
$\ref{prop:distance}$ do not change significantly so long as the
cluster size is roughly uniform and the sampling probability is the
same for all points within a cluster. We will refer to this as the
\emph{clustering assumption}.  In what follows, let $\Pr[s=1 | \C_i ]$
designate the sampling probability for all $x \in \C_i$.  Finally,
define $q(\C_i) = \Pr[s = 1 | \C_i ]$ and $\hat q(\C_i) = |\C_i \cap
S|/|\C_i \cap U|$.
\begin{proposition}
\label{prop:distance}
Let  $B = \displaystyle \max_{i = 1, \ldots, m}
\max(1/q(\C_i), 1/ \hat q(\C_i))$.  Then, the $l_1$ and $l_2$
distances of the distributions $\W$ and $\widehat \W$ can be bounded as
follows,
\begin{equation*}
\small  
{\begin{split}
l_1(\W, \widehat \W) \leq B^2 \sqrt{\frac {|\C_M| k (\log 2k + \log {1
        \over \delta})}{q_0nm}}\ \ l_2(\W, \widehat \W) \leq B^2 \sqrt{\frac {|\C_M| k (\log 2k + \log {1
        \over \delta })}{q_0nm^2}},
\end{split}}
\end{equation*}
where $q_0 = \min q(\C_i)$ and $|\C_M| = \max_i |\C_i|$.
\end{proposition}
\begin{proof}
By definition of the $l_2$ distance,
\begin{equation*}
\small {\begin{split}
l_2^2(\W, \widehat \W)
& = \frac{1}{m^2} \sum_{i=1}^k \sum_{x \in \C_i} \left( {1 \over p(x)} -
{1 \over \hat p(x)} \right)^2 
=  \frac{1}{m^2} \sum_{i=1}^k \sum_{x \in \C_i} \left( {1 \over 
q(\C_i)} - {1 \over \hat q(\C_i)} \right)^2 \\
& \leq \frac{B^4 |\C_M|}{m^2} \sum_{i=1}^k \max_i (q(\C_i) - \hat
q(\C_i))^2.
\end{split}}
\end{equation*}
The right-hand side of the first line follows from the clustering
assumption and the inequality then follows from exactly the same steps as
in Proposition~\ref{prop:distance} and factoring away the sum over the
elements of $\C_i$. Finally, it is easy to see that the $\max_i
(q(\C_i) - \hat q(\C_i))$ term can be bounded just as in
Lemma~\ref{lem:emp_convergence} using a uniform convergence bound, however
now the union bound is taken over the clusters rather than unique points. \qed
\end{proof}
Note that when the cluster size is uniform, then $|C_M| k = m$, and
the bound above leads to an expression similar to that of
Proposition~$\ref{prop:distance}$.  \ignore{Notice however, it is
  always the case that $q_0 \geq p_0$, which is the prime motivation
  for clustering.}

We used the leaves of a decision tree to define the clusters.  A
decision tree selects binary cuts on the coordinates of $x \in X$ that
greedily minimize a node impurity measure, e.g., MSE for regression
\cite{CART}. Points with similar features and labels are clustered
together in this way with the assumption that these will also have
similar sampling probabilities.

\begin{table*}[t]
  \caption{Normalized mean-squared error (NMSE) for various regression data sets using
    unweighted, ideal, clustered and kernel-mean-matched training sample 
    reweightings.}
\label{regression-results}
\vskip -0.17in
\begin{center}
\begin{small}
\begin{sc}
\begin{tabular}{l|lll|llll}
\hline
%\abovespace\belowspace
Data set & $|U|$ & $|S|$ & $n_{test}$ & Unweighted & Ideal & Clustered & KMM \\
\hline
%\abovespace
abalone & 2000 & 724 & 2177 & 0.654$\pm$0.019 & 0.551$\pm$0.032 & 0.623$\pm$0.034 & 0.709$\pm$0.122 \\
bank32nh & 4500 & 2384 & 3693 & 0.903$\pm$0.022 & 0.610$\pm$0.044 & 0.635$\pm$0.046 & 0.691$\pm$0.055 \\
bank8FM & 4499 & 1998 & 3693 & 0.085$\pm$0.003 & 0.058$\pm$0.001 & 0.068$\pm$0.002 & 0.079$\pm$0.013 \\
cal-housing & 16512 & 9511 & 4128 & 0.395$\pm$0.010 & 0.360$\pm$0.009 & 0.375$\pm$0.010 & 0.595$\pm$0.054 \\
cpu-act & 4000 & 2400 & 4192 & 0.673$\pm$0.014 & 0.523$\pm$0.080 & 0.568$\pm$0.018 & 0.518$\pm$0.237 \\
cpu-small & 4000 & 2368 & 4192 & 0.682$\pm$0.053 & 0.477$\pm$0.097 & 0.408$\pm$0.071 & 0.531$\pm$0.280 \\
housing & 300 & 116 & 206 & 0.509$\pm$0.049 & 0.390$\pm$0.053 & 0.482$\pm$0.042 & 0.469$\pm$0.148 \\
kin8nm & 5000 & 2510 & 3192 & 0.594$\pm$0.008 & 0.523$\pm$0.045 & 0.574$\pm$0.018 & 0.704$\pm$0.068 \\
%\belowspace
puma8NH & 4499 & 2246 & 3693 & 0.685$\pm$0.013 & 0.674$\pm$0.019 & 0.641$\pm$0.012 & 0.903$\pm$0.059 \\
\hline
\end{tabular}
\end{sc}
\end{small}
\end{center}
\vskip -.28in
\end{table*}

Several methods for bias correction are compared in
Table~\ref{regression-results}. Each method assigns 
corrective weights to the training samples.  The {\em unweighted}
method uses weight $1$ for every training instance. The {\em ideal}
method uses weight $\frac{1}{\Pr[s = 1 | x]}$, which is optimal but
requires the sampling distribution to be known.  The {\em clustered}
method uses weight $|\C_i \cap U|/|\C_i \cap S|$, where the clusters
$\C_i$ are regression tree leaves with a minimum count of 4
%\ignore{
(larger cluster sizes showed
similar, though declining, performance).
%}
The KMM method uses the approach of \emcite{huang-nips06} with a
Gaussian kernel and parameters $\sigma = \sqrt{d/2}$ for $x \in
\Rset^d$, $B = 1000$, $\epsilon=0$. Note that we know of no
principled way to do cross-validation with KMM since it cannot
produce weights for a held-out set \cite{sugiyama-nips2008}.

The regression datasets are from LIAAD\footnote{{\tt \small
    www.liaad.up.pt/\~{}ltorgo/Regression/DataSets.html.}} and are
sampled with $P[s = 1 | x] = \frac{e^v}{1 + e^v}$ where $v = \frac{4 w
  \cdot (x - \bar{x})}{\sigma_{w \cdot (x - \bar{x})}}$, $x \in
\Rset^d$ and $w \in \Rset^d$ chosen at random from $[-1,1]^d$. In our
experiments, we chose ten random projections $w$ and reported results
with the $w$, for each data set, that maximizes the difference between
the unweighted and ideal methods over repeated sampling trials. In
this way, we selected bias samplings that are good candidates for bias
correction estimation.

For our experiments, we used a version of SVR available from
LibSVM\footnote{{\tt \small
    www.csie.ntu.edu.tw/\~{}cjlin/libsvmtools.}} that can take as
input weighted samples, with parameter values $C = 1$, and $\epsilon =
0.1$ combined with a Gaussian kernel with parameter $\sigma =
\sqrt{d/2}$. We report results using normalized mean-squared error
(NMSE): $\frac{1}{n_{test}} \sum^{n_{test}}_{i=1} \frac{(y_i -
  \hat{y}_i)^2}{\sigma_y^2}$, and provide mean and standard deviations
for ten-fold cross-validation.

Our results show that reweighting with more reliable counts, due to
clustering, can be effective in the problem of sample bias
correction. These results also confirm the dependence that our
theoretical bounds exhibit on the quantity $n_0$. The results obtained
using KMM seem to be consistent with those reported by
the authors of this technique.\footnote{We thank Arthur Gretton for
discussion and help in clarifying the choice of the parameters and 
design of the KMM experiments reported in \cite{huang-nips06}, and for
providing the code used by the authors for comparison studies.}

\section{Conclusion}
\label{sec:Conclusion}

We presented a general analysis of sample selection bias correction
and gave bounds analyzing the effect of an estimation error on the
accuracy of the hypotheses returned. The notion of distributional
stability and the techniques presented are general and can be of
independent interest for the analysis of learning algorithms in other
settings. In particular, these techniques apply similarly to other
importance weighting algorithms and can be used in other contexts such
that of learning in the presence of uncertain labels.
The analysis of the discriminative method of \cite{bickel-icml07} for
the problem of covariate shift could perhaps also benefit from this
study.

\bibliographystyle{mlapa}
\bibliography{bias}

\appendix

\section{Proof of Theorem \ref{th:kmm}}
\begin{proof}
  Assume that $\mu(p) = \mu(q)$ for two probability distributions $p$
  and $q$ in $\P$. It is known that if $\E_{x \sim p} [f(x)] = \E_{x
    \sim q} [f(x)]$ for any $f \in C(X)$, then $p = q$. Let $f \in
  C(X)$ and fix $\e > 0$.  Since $K$ is universal, there exists a function
  $g$ induced by $K$ such that $\|f - g\|_\infty \leq \e$.
$\E_{x \sim p} [f(x)] - \E_{x \sim q} [f(x)]$ can be rewritten as
\begin{equation}
\E_{x \sim p} [f(x) - g(x)] + 
\E_{x \sim p} [g(x)] - \E_{x \sim q} [g(x)] + 
\E_{x \sim q} [g(x) - f(x)].
\end{equation}
Since $\bigl|\E_{x \sim p} [f(x) - g(x)]\bigr| \leq \E_{x \sim p}
|f(x) - g(x)| \leq \|f - g\|_\infty \leq \e$ and similarly
$\bigl|\E_{x \sim q} [f(x) - g(x)]\bigr| \leq \e$,
\begin{equation}
\label{ineq1}
\biggl|\E_{x \sim p} [f(x)] - \E_{x \sim q} [f(x)]\biggr| \leq
\biggl|\E_{x \sim p} [g(x)] - \E_{x \sim q} [g(x)] \biggr| + 2 \e.
\end{equation}
Since $g$ is induced by $K$, there exists $w \in F$ such that for all
$x \in X$, $g(x) = \iprod{w}{\Phi(x)}$. Since $F$ is separable, it
admits a countable orthonormal basis $(e_n)_{n \in \Nset}$. For $n
\in \Nset$, let $w_n = \iprod{w}{e_n}$ and $\Phi_n(x) =
\iprod{\Phi(x)}{e_n}$. Then, $g(x) = \sum_{n = 0}^\infty w_n
\Phi_n(x)$. For each $N \in \Nset$, consider the partial sum $g_N(x) =
\sum_{n = 0}^N w_n \Phi_n(x)$. By the Cauchy-Schwarz inequality,
\begin{equation}
|g_N(x)| \leq \|\sum_{n = 0}^N w_n e_n\|_2^{1/2} \|\sum_{n = 0}^N
\Phi_n(x) e_n\|_2^{1/2} \leq \|w\|_2^{1/2} \|\Phi(x)\|_2^{1/2}.
\end{equation}
Since $K$ is universal, it is continuous and thus $\Phi$ is also
continuous \cite{steinwart-jmlr02}. Thus $x \mapsto
\|\Phi(x)\|_2$ is a continuous function over the compact $X$ and
admits an upper bound $B \geq 0$. Thus, $|g_N(x)| \leq
\sqrt{\|w\|_2 B}$.  The integral $\int \bigl|\sqrt{\|w\|_2 B}\bigl| dp$
is clearly well defined and equals $\sqrt{\|w\|_2 B}$. Thus, by the 
Lebesgue dominated convergence theorem, the following holds:
\begin{equation}
\E_{x \sim p} [g(x)] = \int \sum_{n = 0}^\infty w_n \Phi_n(x) dp(x) =
\sum_{n = 0}^\infty w_n \int \Phi_n(x) dp(x).
\end{equation}
By definition of $\E_{x \sim p}[\Phi(x)]$, the last term is
the inner product of $w$ and that term. Thus, 
\begin{equation}
\E_{x \sim p} [g(x)] = \iprod{w}{\E_{x \sim p}\bigl[\Phi(x)\bigr]} = \iprod{w}{\mu(p)}.
\end{equation}
A similar equality holds with the distribution $q$, thus,
\begin{equation*}
\E_{x \sim p} [g(x)] - \E_{x \sim q} [g(x)] = \iprod{w}{\mu(p) - \mu(q)} = 0.
\end{equation*}
Thus, Inequality~\ref{ineq1} can be rewritten as 
\begin{equation}
\biggl|\E_{x \sim p} [f(x)] - \E_{x \sim q} [f(x)]\biggr| \leq 2 \e,
\end{equation}
for all $\e > 0$. This implies $\E_{x \sim p} [f(x)] = \E_{x \sim q} [f(x)]$
for all $f \in C(X)$ and the injectivity of $\mu$.\qed
\end{proof}

\end{document}